\documentclass[journal]{IEEEtran}

\usepackage{amsmath,amssymb,amsfonts}
\usepackage{graphicx}
\usepackage{pifont}

\usepackage{amsthm}
\usepackage{xcolor}
\usepackage[figuresright]{rotating}

\usepackage{graphicx}
\graphicspath{{/}{fig/}}

\usepackage{array}
\usepackage{textcomp}
\usepackage{xcolor}
\usepackage{multirow}
\usepackage{booktabs}

\usepackage{mathtools}
\usepackage{breqn}
\usepackage{float}

\usepackage{pgfplots}
\pgfplotsset{compat=1.7}
\usepgfplotslibrary{groupplots}

\usepackage{caption}
\usepackage{subcaption}

\usepackage{multirow,tabularx}
\usepackage{hyperref}
\usepackage[vlined, ruled, shortend]{algorithm2e}
\usepackage{enumerate}

\usepackage{listings, xcolor}

\definecolor{verylightgray}{rgb}{.97,.97,.97}

\lstdefinelanguage{Solidity}{
	keywords=[1]{anonymous, assembly, assert, balance, break, call, callcode, case, catch, class, constant, continue, constructor, contract, debugger, default, delegatecall, delete, do, else, emit, event, experimental, export, external, false, finally, for, function, gas, if, implements, import, in, indexed, instanceof, interface, internal, is, length, library, log0, log1, log2, log3, log4, memory, modifier, new, payable, pragma, private, protected, public, pure, push, require, return, returns, revert, selfdestruct, send, solidity, storage, struct, suicide, super, switch, then, this, throw, transfer, true, try, typeof, using, value, view, while, with, addmod, ecrecover, keccak256, mulmod, ripemd160, sha256, sha3}, 
	keywordstyle=[1]\color{blue}\bfseries,
	keywords=[2]{address, bool, byte, bytes, bytes1, bytes2, bytes3, bytes4, bytes5, bytes6, bytes7, bytes8, bytes9, bytes10, bytes11, bytes12, bytes13, bytes14, bytes15, bytes16, bytes17, bytes18, bytes19, bytes20, bytes21, bytes22, bytes23, bytes24, bytes25, bytes26, bytes27, bytes28, bytes29, bytes30, bytes31, bytes32, enum, int, int8, int16, int24, int32, int40, int48, int56, int64, int72, int80, int88, int96, int104, int112, int120, int128, int136, int144, int152, int160, int168, int176, int184, int192, int200, int208, int216, int224, int232, int240, int248, int256, mapping, string, uint, uint8, uint16, uint24, uint32, uint40, uint48, uint56, uint64, uint72, uint80, uint88, uint96, uint104, uint112, uint120, uint128, uint136, uint144, uint152, uint160, uint168, uint176, uint184, uint192, uint200, uint208, uint216, uint224, uint232, uint240, uint248, uint256, var, void, ether, finney, szabo, wei, days, hours, minutes, seconds, weeks, years},	
	keywordstyle=[2]\color{teal}\bfseries,
	keywords=[3]{block, blockhash, coinbase, difficulty, gaslimit, number, timestamp, msg, data, gas, sender, sig, value, now, tx, gasprice, origin},	
	keywordstyle=[3]\color{violet}\bfseries,
	identifierstyle=\color{black},
	sensitive=false,
	comment=[l]{//},
	morecomment=[s]{/*}{*/},
	commentstyle=\color{gray}\ttfamily,
	stringstyle=\color{red}\ttfamily,
	morestring=[b]',
	morestring=[b]"
}

\newlength\figureheight
\newlength\figurewidth
\SetAlCapNameFnt{\footnotesize}
\SetAlCapFnt{\footnotesize}

\title{
    Partition-Tolerant and Byzantine-Tolerant Decision-Making for Distributed Robotic Systems with IOTA and ROS\,2
}

\author{Farhad Keramat,
        Jorge Peña Queralta,
        and~Tomi Westerlund
    \thanks{Farhad Keramat, Jorge Peña Queralta and Tomi Westerlund are with the \href{https://tiers.utu.fi}{Turku Intelligent Embedded and Robotic Systems (TIERS) Lab, University of Turku, Turku, Finland}, e-mails: \{fakera, jopequ, tovewe\}@utu.fi.}
}

\begin{document}

\maketitle

\begin{abstract}%
    \label{sec:abstract}%
    With the increasing ubiquity of autonomous robotic solutions, the interest in their connectivity and in the cooperation within multi-robot systems is rising. Two aspects that are a matter of current research are robot security and secure multi-robot collaboration robust to byzantine agents. Blockchain and other distributed ledger technologies (DLTs) have been proposed to address the challenges in both domains. Nonetheless, some key challenges include scalability and deployment within real-world networks. This paper presents an approach to integrating IOTA and ROS\,2 for more scalable DLT-based robotic systems while allowing for network partition tolerance after deployment. This is, to the best of our knowledge, the first implementation of IOTA smart contracts for robotic systems, and the first integrated design with ROS\,2. This is in comparison to the vast majority of the literature which relies on Ethereum. We present a general IOTA+ROS\,2 architecture leading to partition-tolerant decision-making processes that also inherit byzantine tolerance properties from the embedded blockchain structures. We demonstrate the effectiveness of the proposed framework for a cooperative mapping application in a system with intermittent network connectivity. We show both superior performance with respect to Ethereum in the presence of network partitions, and a low impact in terms of computational resource utilization. These results open the path for wider integration of blockchain solutions in distributed robotic systems with less stringent connectivity and computational requirements.
\end{abstract}

\IEEEpeerreviewmaketitle


\section{Introduction}\label{sec:introduction}

Autonomous robots are revolutionizing industries and civil applications. Two aspects that are part of today's ubiquitous robotic solutions are connectivity and teaming~\cite{schranz2020swarm, javaid2021substantial, queralta2020collaborative}. Indeed, many robots today are deployed as part of larger fleets or in teams, often heterogeneous and fruit of the combination of robots from different vendors.
In addition, the proliferation of robotic systems is leading to increased connectivity and reliance on multi-robot interaction or cloud-based services~\cite{saha2018comprehensive}. Maintaining security becomes more critical as robots become more connected and ubiquitous. Additionally, multi-robot systems may be prone to malicious behavior from within the system, 
while malfunctioning units or sensors can also lead to unexpected functionality or results. In practical applications, a key aspect and important part of a multi-robot systems is collective decision-making~\cite{brambilla2013swarm}, which is highly susceptible to malicious behavior. However, the potential effects of byzantine agents has not been always considered in the literature~\cite{olfati2004consensus}.

In general, as autonomous robotic solutions become more widely deployed, more attention is put to connectivity and management of larger-scale distributed systems. With increased connectivity, however, also comes an increased risk of cybersecurity threats~\cite{white2021usable}. A robot featuring different forms or wireless connectivity opens the door to a number of attack vectors. Indeed, recent research has shown that multiple commercial platforms are susceptible to hijacking from different types of interfaces~\cite{mayoral2022robot}. With robots being cyber-physical systems that often interact with humans and their environment, a security vulnerability becomes both a safety and security risk. In this work, we focus on securing multi-robot interaction and collaborative decision-making from the perspective of a distributed networked system where data is shared and collaborative decisions are made.

\begin{figure}
    \centering
    \includegraphics[width=0.42\textwidth]{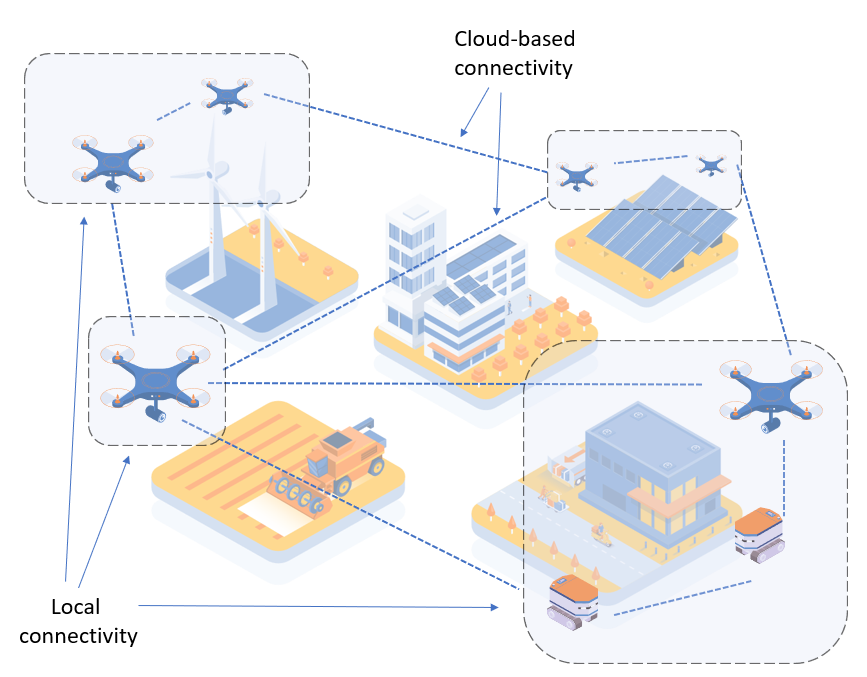}
    \caption{Large-scale deployments of connected robots often include both cloud-based connectivity and local connectivity within subsets of robots, leading to potential network partitions. Global peer-to-peer connection cannot be always relied upon when deployments occur in remote areas. This is particularly critical when considering traditional blockchain-based solutions in which data is lost if a subset of robots are disconnected for a period of time.}
    \label{fig:concept}
\end{figure}

The advances in distributed ledger technologies (DLTs), which offer consensus mechanisms among multiple untrusted parties, have made them widely used in interconnected network of devices in recent years. In addition to providing a consensus mechanism, DLT integration also introduces identity management, tamper-proof logging, and smart contract execution, all of which can benefit distributed robotic systems~\cite{queralta2020enhancing}. The majority of solutions for the IoT and robotic applications in the literature rely on Ethereum smart contracts~\cite{nawaz2020edge, queralta2021blockchain}. However, there are limitations in terms of throughput and tolerance against network partitioning. Network partitioning is one of the challenges in the integration of DLTs in distributed and mobile robotic systems due to the highly dynamic network topologies and limited bandwidths available. Multiple works in the literature already show the potential of blockchain technology for managing byzantine behaviour or consensus in swarms of robots~\cite{strobel2020blockchain, ferrer2021following}. In contrast to the majority of Ethereum-based systems, IOTA has already been identified as a solution to address these issues~\cite{santos2021towards}, but smart contracts have not been available until very recently.

In this paper, we propose a methodology to achieve collective decision-making in distributed robotic applications in a partition-tolerant manner. We achieve this by utilizing IOTA smart contract platform. This methodology inherently integrates, as well, byzantine-tolerant processes. In addition, we propose a novel architecture to integrate IOTA's two-layer structure with ROS\,2.To the best of our knowledge, this is the first approach to integrate IOTA's smart contracts with ROS\,2 to leverage DLT in multi-robot systems. We chose a distributed collaborative mapping task to demonstrate how the proposed methodology can be applied on top of our architecture. The distributed collaborative mapping task is simulated in larger-scale in Gazebo as well as with a real-world experiment. The implementations are open-source as the first integration of ROS\,2 with IOTA's smart contract. 

In summary, the main contributions of this work are the following: 
\begin{enumerate}[(i)]
    \item the first general methodology for DLT-based robotic applications to achieve partition-tolerant collective decision-making, which also inherits byzantine tolerance from the blockchain,
    \item an architectural design for integrating IOTA and ROS\,2 for distributed multi-robot systems, and
    \item a demonstration of the applicability of the proposed methodology and architecture for a multi-robot collaborative mapping application.
\end{enumerate}

The rest of the paper is organized as follows. Section~\ref{sec:related_work} introduces previous research in the collective decision-making problems and the use of distributed ledger technologies for robotic systems and introduces the key concepts behind smart contracts in both Ethereum and IOTA. Then, in Section~\ref{sec:methods} we describe the methodology to make partition-tolerant decision-making protocols and how to design approaches for integrating IOTA and ROS\,2 into the same framework. A partition-tolerant fault-tolerant distributed collaborative mapping algorithm is shown in Section~\ref{sec:results}, with a comparison between Ethereum and IOTA. A discussion on scalability and future potential appears in Section~\ref{sec:discussion}. Finally, Section~\ref{sec:conclusion} concludes the work and lays out the directions for future work.


\section{Related Works}
\label{sec:related_work}

This section briefly reviews the literature in robot cybersecurity, and the potential of blockchain and other distributed ledger technologies for securing and building trust in robot swarms.

Multiple research efforts have been directed towards securing robotic systems. For example, in~\cite{clark2017cybersecurity}, Clark et al. study security threats on robots at the hardware, firmware, and application layer and lists possible attacks from spoofing sensor data to denial of service attacks. In~\cite{yaacoub2021robotics}, the focus is on the impact of the security attacks and suggests some countermeasures. In another work, Higgins et al. look at both robotics and security perspectives~\cite{higgins2009threats}. In this study, the the authors compare swarm robotics use cases to similar technologies in order to find the unique features making similar security measurements unfeasible or ineffective. In general, the security of swarm robotics is very crucial in defence, healthcare, environmental, and commercial applications~\cite{higgins2009survey}.

Collective decision-making is an essential element in swarm robotics and have studied extensively. In these studies, all robots in the swarm are assumed to be honest and protocol obedient~\cite{millard2013towards}. But a single intruder robot can easily affect the entire system's decision-making. For this reason, recent studies have also taken into account probable faulty robots to make their approach resilient against them. Sargeant and Tomlinson~\cite{sargeant2013modelling} give a generic swarm model and how a malicious intruder can be modeled in this context. In \cite{zikratov2016dynamic}, Zikratov et al. propose a dynamic trust management framework that enables robots in an ad-hoc network to detect a compromised device and an access control unit that expects newly joined members to behave honestly up to a certain time to participate in decision-making tasks. For multi-robot systems with time-varying communication graphs dealing with malicious parties is more challenging. In~\cite{saldana2017resilient}, a consensus approach is proposed, resilient if communication graphs reunite in a bounded time period. The method proposed in this work, is a more general approach to solve this kind of issues.

Despite the recent research efforts to design and develop decision-making methods that are fault-tolerant, the literature contains mainly specialized approaches for specific use cases or applications scenarios. To the best of our knowledge, there is a lack of a generic and scalable collaborative decision-making framework for distributed robotic systems. Within this domain, an early work by Castell\'o Ferrer pointed at the applicability of blockchain technologies in swarm robotics~\cite{castello2018blockchain}. The study discusses how this technology benefits security, distributed decision-making, and new business models in robotics. The work also points out that applying blockchain to resource constraint devices (e.g., mobile robots) can be challenging. However, blockchain technology is still presented as the potential infrastructure to ensure security and safety regulations for robotics. In a more recent work, Afanasyev et al. list open issues in combining blockchain and robotics and certain application scenarios~\cite{afanasyev2019blockchain}. One of the benefits of blockchain technology is that robots can be assigned specific tasks through smart contracts. In our study, we present smart contracts as the backbone of collaborative decision-making. A proof of concept showcasing blockchain in robotic swarms was demonstrated by Strobel et al. in~\cite{strobel2020blockchain}. In the experiments in~\cite{strobel2020blockchain}, Robots in a swarm collaboratively reach an agreement on the most common tile color in an environment with white and black tiles. As opposed to conventional methods, the proposed blockchain-based solution on Ethereum~\cite{wood2014ethereum} can tolerate malicious behavior. Another example of collective decision-making in swarm robotics in~\cite{ferrer2021following} presents an in-depth study of following-the-leader problems. In swarm robotics literature, Ethereum is the most commonly used blockchain platform. However, it comes with limitations in terms of scalability and deployment in embedded systems. Other platforms have emerged that potentially solve some of these issues, such as Hyperledger Fabric~\cite{androulaki2018hyperledger} and IOTA~\cite{popov2018tangle}. These have already been explored in some works, albeit more limited experiments have been demonstrated~\cite{salimi2022towards, santos2021towards, salimi2022secure}.

From a system design perspective, blockchain technology brings benefits in terms of immutability of past data, and distributed decision making, among others. However, it also brings new challenges. In most existing use cases, blockchains are deployed between nodes (e.g., computers or servers) with a stable network connection. This assumption, nonetheless, does not necessarily hold when operating swarms of mobile robots. To address the issue of potentially intermittent connectivity, Tran et al. proposed SwarmDAG~\cite{tran2019swarmdag}. SwarmDAG is a system-level design based on directed acyclic graphs (DAGs) that incorporates a membership management system to handle new members. Early solutions like SwarmDAG, however, become vulnerable to security issues that traditional blockchains already solve (e.g., Sybil attacks). Next-generation blockchain systems that are intrinsically based on DAGs, such as IOTA, are able to provide both scalability and security. For example, a surveillance system is presented in~\cite{santos2021towards} that also tolerates partitioning within the network while maintaining secure consensus with IOTA. The research in~\cite{santos2021towards} was carried out before smart contracts were developed for IOTA. The IOTA foundation has now introduced IOTA smart contracts that run on chains over the core DAG structure. The technology is therefore now ready for more complex designs and the integration of distributed decision-making processes, taking advantage of smart contracts and asynchronous calls in IOTA. Compared to the state-of-the-art in blockchain-based robotic systems, we propose in this paper a novel approach with a general methodology for making virtually any decision-making problem in a distributed and partition-tolerant manner, given that it can be implemented as a smart contract with a series of constraints. This allows for byzantine-tolerant consensus in multi-robot systems and robot swarms without strong connectivity requirements. At the same time, this solution achieves higher degrees of scalability when compared to traditional blockchains that form the vast majority of the work to date in DLTs within the robotics field.


\begin{figure}
    \centering
    \includegraphics[width=.48\textwidth]{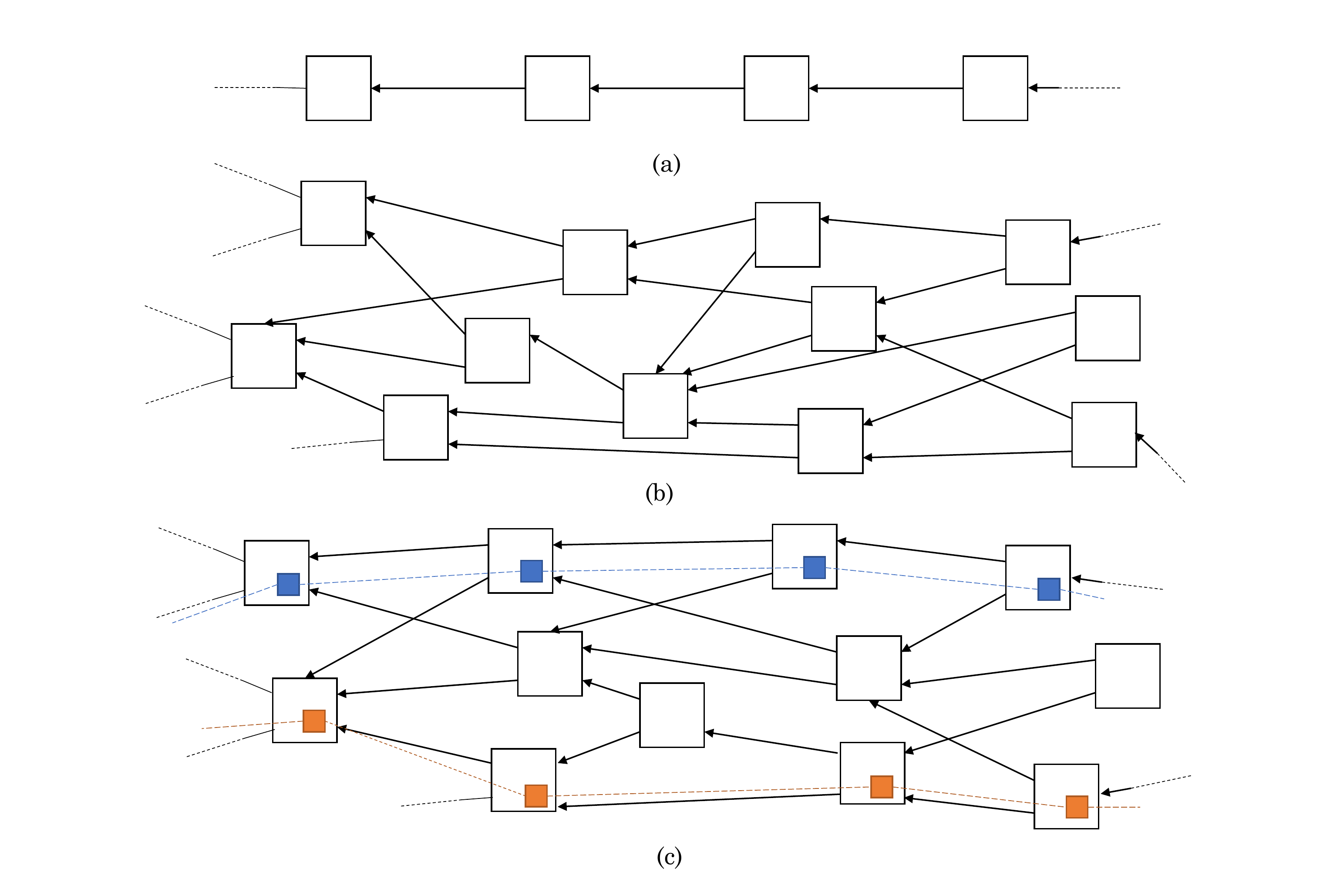}
    \caption{Illustration of (a) a traditional blockchain, (b) the IOTA Tangle, and (c) the two-layer ISCP architecture where we highlight two possible chains anchored to different Table nodes.}
    \label{fig:concept_architectures}
\end{figure}

\section{Background} 

Blockchain systems, a subset of the wider domain of DLTs, have grown in popularity over the past few years, partly owing to the public interest around cryptocurrencies. Bitcoin~\cite{nakamoto2008bitcoin} was the first cryptocurrency that removed the trust on a third party to conduct a transaction between two entities. Although financial transactions were the initial objective, blockchains can be used in a variety of use cases. In addition to enabling distributed decision-making, as we discussed earlier, blockchains also serve as a tamper-proof log. Blockchain technology has been used in some robotics research just for logging purposes to have a reliable history of events~\cite{lopes2019robot}. Smart contracts, which are computer programs automatically executed when a set of conditions is fulfilled, have opened up new possibilities for distributed applications (DApps). In the following, we describe the key concepts behind the more traditional Ethereum blockchain and the DAG-based IOTA architecture. These different design approaches are illustrated in Fig.~\ref{fig:concept_architectures}.

\textbf{Ethereum}:    
    Ethereum made a significant impact on distributed ledger technologies by introducing Turing-complete smart contracts. Solidity is Ethereum's programming language exclusively designed for coding smart contracts. A Turing-complete smart contract has enabled Ethereum to be widely adopted in a wide range of domains. Researchers have also exploited Ethereum's capabilities within multi-robot systems~\cite{strobel2020blockchain}. While Ethereum introduced new possibilities to blockchain systems, it still uses the classical single-chain structure. Therefore, the intrinsic scalability problem remains. In addition to scalability issues, and considering the perspective of the design of multi-robot systems, tolerating network partitioning is imperative for real-world deployments in many application scenarios. The Ethereum foundation is working on Ethereum 2.0, which is posed to resolve many of the current scalability issues by leveraging \textit{sharding} (a new approach to achieving consensus within subsets of the global network); however, the solutions are not matured yet.
    
\textbf{Tangle (IOTA)}:
    The Tangle, a DAG-based DLT, was introduced by S. Popov. to solve some of the underlying deficiencies in classic blockchain systems~\cite{popov2018tangle}. The Tangle is the underlying structure used by the IOTA DLT. Most of the blockchain systems to date use blocks encapsulating a set of transactions as their primary data structure. These data blocks are then usually connected as a linked list by using the hash of each block as the linking element between consecutive entries in the list. In the Tangle, transactions themselves are the primary data structures. Using individual transactions as the primary data structure enables even nodes with limited resources to participate in the consensus. The DAG structure is then generated as each transaction must refer to two previous unconfirmed transactions based on the view of the Tangle that the node that issues it has.
    The core idea behind the Tangle is that keeping the ledger on a graph rather than a single (linear) chain would allow for a certain level of flexibility in terms of network partitioning. This makes IOTA, a priori, an excellent DLT solution for multi-robot systems~\cite{tran2019swarmdag, santos2021towards}.
    
\textbf{Shimmer (IOTA 2.0)}:
    The first version of IOTA focused on making the graph-based data structure functional while preserving the security standards of most DLT solutions. In this version, only basic transactions were possible (e.g., financial transactions, such as a exchange of tokens, or data publishing for IoT devices). To keep the Tangle stable and secure, a centralized Coordinator managed by the IOTA Foundation was in charge of confirming valid transactions.
    To achieve full decentralization, the IOTA Foundation redesigned the Tangle and launched Shimmer as the second version of IOTA. GoShimmer is the Go implementation of Shimmer clients, which will be used in this work. 

\textbf{Wasp (IOTA Smart Contracts)}:
    Due to the graph-based data structure in the Tangle, embedding a smart contract mechanism in IOTA was a bigger challenge than in traditional blockchain. In general terms, smart contracts are made possible in a blockchain through a state machine that has a state that can be altered by entering a new block. Such state machine requires a global state and is therefore not directly \textit{embeddable} within the Tangle. To solve this issue, the IOTA Foundation uses the Tangle as a first layer, on top of which they introduce the IOTA Smart Contract Platform (ISCP) as a second layer. Wasp refers to the Go implementation of the ISCP client. ISCP clients or Wasp nodes can create a chain in this second layer. Each of these chains can be compared to an Ethereum blockchain, in this case having every block \textit{anchored} to the first layer. Other Wasp nodes can join the chain, and all the Wasp nodes participating in a chain form a \textit{chain committee}. Similar to an Ethereum blockchain, committee members can run smart contracts in the chains they belong to. Each \textit{chain committee} has a finite number of members, meaning they can run a byzantine fault tolerant (BFT) algorithm to reach consensus. BFT consensus algorithms can tolerate at most one-third of byzantine members. It is possible to make virtually any number of chains on top of the first layer with different committees. In addition, these chains can interact with each other, referred to as asynchronous calls in ISCP. Asynchronous calls enable smart contracts to call a method of another smart contract in a different chain. It is worth mentioning that, while global connectivity is not required at all times within the Tangle, running a smart contract requires consensus to be reached by members of the corresponding Wasp committee. Therefore, at least two-thirds of the nodes participating in a chain need to be in a common network partition when chain blocks are committed. This requirement does not extend to two chains interacting through asynchronous calls except when the calls occur.

\textbf{Robotics Middleware:} the Robot Operating System (ROS) is the de-facto standard in today's autonomous robots~\cite{quigley2009ros}. From the perspective of multi-robot systems and distributed networked systems, the original ROS\,1 version has certain limitations, mainly due to the existence of a certain node managing the interaction between the different actors in the system. The new ROS\,2 version solves this with the introduction of the data distribution service (DDS) standard for the lower-level communication middleware. DDS also provides a security extension empowering ROS\,2 itself. Only ROS\,2 is a natural selection for integration with distributed ledger technologies. In addition to the DLT platform in use, the distributed communication and security that DDS enable are crucial features for a secure and trustable system. This work does not aim at replacing those features, but instead complimenting them with an additional channel for building trust and implementing collaborative decision-making processes. ROS\,2 already provides tools for data encryption and data access control.

In summary, to address the network partitioning problem in multi-robot systems, and enable dynamic network topologies, IOTA is a promising solution. This has already been showcased with a proof of concept in the literature~\cite{santos2021towards}, however, with the lack of smart contracts for more complex integrations. With an IOTA-based system, disconnected robots or separate groups of robots can still operate on the same global Tangle, with the corresponding transactions in separate subgraphs that are merged whenever global connectivity is regained. With the introduction of ISCP and the asynchronous calls, in addition to this, we can also design systems that run distributed applications inherently able to tolerate network partitioning. In this paper, we propose a general approach for making distributed, partition-tolerant decision-making tasks through IOTA smart contracts. We design such approach to be integrated with ROS\,2 in a seamless way. This integration is two-directional, with ROS\,2 feeding data to the IOTA Tangle and IOTA smart contracts being used to implement functionality that replaces distributed ROS\,2 nodes. In particular, we exploit the asynchronous calls for \textit{connecting} chains that \textit{live} in different network partitions.


\begin{figure*}[t]
    \centering
    \includegraphics[width=\textwidth]{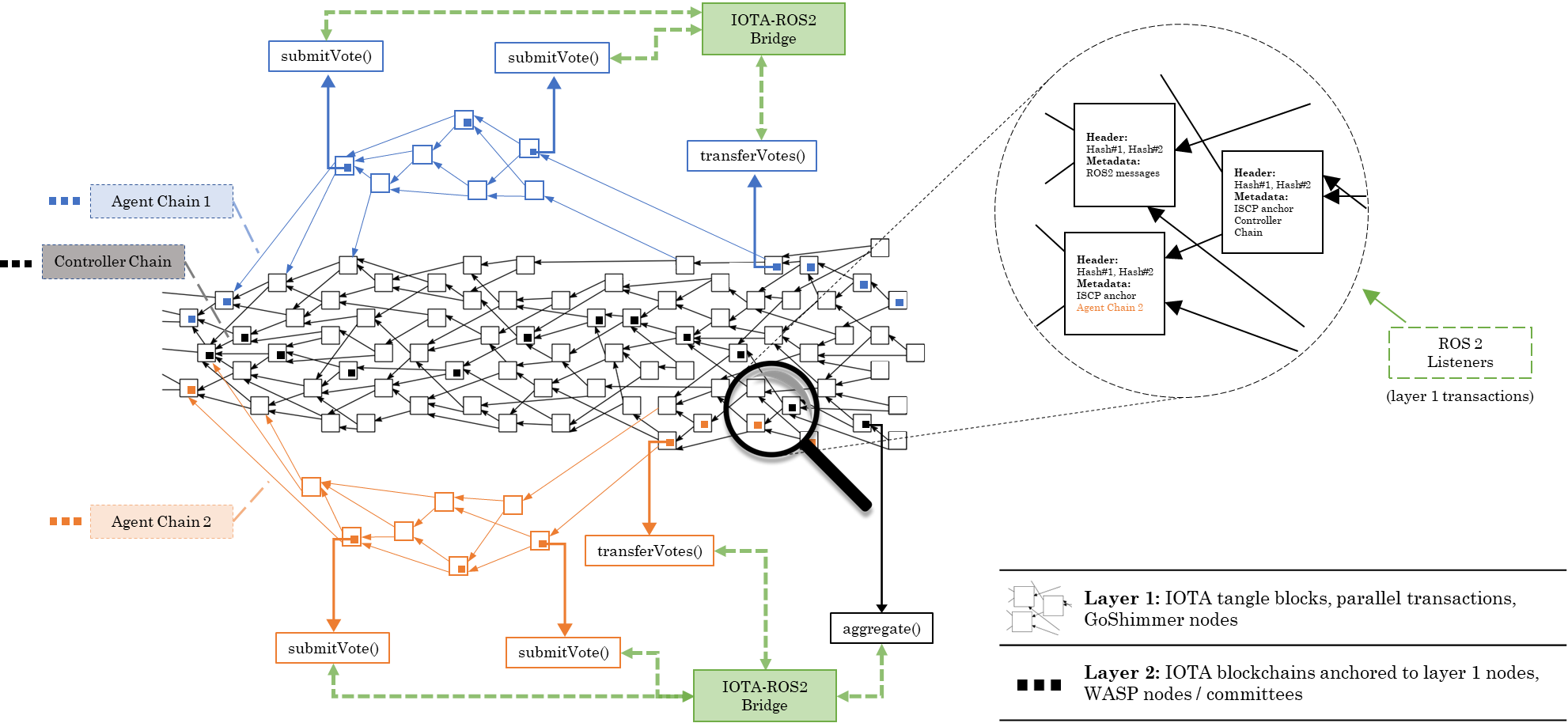}
    \caption{Illustration of the integration of ROS\,2 with the IOTA tangle (layer 1) and IOTA smart contracts (layer 2) enabling secure, distributed and partition-tolerant decision-making. The main network partition, represented in black, includes the core \textsc{Controller Swarm} chain as well as other nodes with global connectivity. In orange and blue we illustrate two examples of transactions (both layer-1 and layer-2 chains) related to mobile robots that have intermittent connectivity and thus create their own network partitions. These partitions can also be formed by groups of robots that remain locally connected but globally disconnected to the rest of the network. Both layer-1 and layer-2 nodes are bridged to ROS\,2, with layer-1 nodes being passive listeners that dump ROS\,2 data to the tangle. Layer-2 nodes have full bi-directional communication between IOTA and ROS\,2 and implement the integration interface.}
    \label{fig:iotaarc}
\end{figure*}

\section{Methodology}\label{sec:methods}

This section covers the overall system design, including how robots operate with the two-layered IOTA architecture and how IOTA is bridged with ROS\,2. The section focuses on how smart contracts are designed and deployed to allow for network partitions and intermittent connectivity while maintaining global consensus at the end of a distributed multi-robot mission. 

\subsection{Problem Definition}

Collaborative decision-making processes in multi robot systems often refer to any method that requires the combination of input, or data, from different robots and processes the data or makes a decision based on it in a decentralized manner. Examples include distributed role allocation algorithms, distributed perception, cooperative mapping, or decentralized formation control, among others. We assume in all cases that robots operate as individual entities and that the decision-making process is not governed by a central authority. These processes are used to achieve agreement and specialization in multi-robot systems~\cite{brambilla2013swarm}. In this work, we focus only on reaching agreements on a series or variables, often referred to as consensus problems.

The core objective of this work is to provide a framework for collaborative decision-making in a multi-robot system based on IOTA. To this end, we integrate IOTA smart contracts with ROS\,2 nodes, in a way that data fed to IOTA from ROS\,2 is processed within smart contracts. The results of such process are then fed back to ROS\,2 topics, even though ROS services or actions could also be implemented in the future. To achieve this goal, this section covers the first two contributions of our work:
\begin{enumerate}[i.]
    \item First, we introduce a methodology and architecture for integrating IOTA with ROS\,2 nodes. We explain how ROS\,2 data and processes can be either be part of the IOTA Tangle or be implemented as smart contracts in IOTA's second layer.
    \item Second, we propose an strategy for enabling partition-tolerant decision-making through the combination of multiple smart contracts in different chains, each existing in its own network partition within the Tangle.
\end{enumerate}

In addition to introducing this approach, our experimental results reported in the next section focus on a proof of concept of the proposed methodology with a cooperative mapping use case, covering the third contribution of our work.

\subsection{System Architecture: Integrating IOTA and ROS\,2}

To leverage IOTA's smart contracts in the multi-robot systems, we propose an integration of ROS\,2 with the two-layer IOTA architecture illustrated in Fig.~\ref{fig:iotaarc} as described in the following:

\textbf{Goshimmer network:} Designed for lightweight IoT nodes, every robot in the system can run a shimmer node. A cloud fleet management system, or fixed sensors and infrastructure with connectivity can also be part of the Goshimmer network. In fact, the more nodes that participate in the network, the more secure that IOTA's Tangle is. Therefore, and within the system performance limits, the Goshimmer network is a good place to publish higher frequency transactions (e.g., raw sensor data from the robots that can be also used for validation in different processes later on).

Goshimmer nodes running in the robots are thus also ROS\,2 \textit{listeners}, i.e., they subscribe to a topic and re-publish the data in the Tangle. The nodes can also be configured to publish data from the Tanble (e.g., metadata from other transactions, data about the Tangle topology, or even the node's own real-time performance data). To implement this, we integrate the Goshimmer library with \textit{rclgo}, the ROS\,2 client library for Go. Publishing ROS\,2 messages on the Tangle would generate a tamper-proof log of events.

\textbf{Wasp committees:} Running a Wasp node on every robot in the system is unnecessary; instead, we propose to run Wasp nodes on the robots with higher processing power or the cloud fleet management system. In order to maintain security, Wasp nodes must be distributed evenly throughout the system. We propose that at least $3f+1$ robots with higher processing power run a Wasp node in each subgroup of robots that might have intermittent connectivity to tolerate $f$ malicious robot. 
Every Wasp node should be connected to a Goshimmer node to interact with L1. Since we proposed to run a Goshimmer node on every robot, a Wasp node should connect to the Goshimmer node running on the same robot.

According to applications running on the multi-robot system, Wasp nodes should create chains and form committees. All the robots working on an application can operate on a single chain in case they don't use interchain asynchronous calls. If the application utilizes interchain asynchronous calls, Wasp nodes should be distributed on the different chains accordingly.

\textbf{ROS\,2 bridge:} We have implemented a ROS\,2 listener node with \textit{rclgo} that subscribes to a topic and forwards these messages or hash of the messages to Goshimmer nodes. Goshimmer nodes will use these data to create new transactions in order to keep the network alive. Keeping these messages on the ledger will provide a complete log of events in multi-robot systems. In addition to a ROS\,2 listener node, a ROS\,2 talker node can be implemented to read on-ledger transactions and publish them for log reviewing and validating processes.

For Wasp nodes, a similar approach is utilized to interact with ROS\,2. A Go script is utilized to call methods of smart contracts and also retrieve the results from it. This script consists of a ROS\,2 subscriber which listens on a topic to receive the command to call a method from a smart contract on the chain specified with \textit{ChainID}. If the method has a returning value, the script will retrieve the value and publish it on a predefined topic. Also, ROS\,2 actions could be a better alternative instead of using a listener and talker topics.

Figure.~\ref{fig:iotaarc} illustrates part of the Tangle and the Wasp chain blocks anchored to the Tangle. In the figure, time increases from left to right. Empty squares show L1 transactions created by Goshimmer nodes. In each transaction, the header includes the hash of two referred transactions and a metadata part. This metadata can be dumped ROS\,2 messages or the anchored state of a chain. Filled squares inside an empty square shows a block of the L2 chain created by the Wasp committee. Three chains are shown in this figure, colored blue, orange, and black, with only one smart contract deployed on each chain. Methods called from each smart contract are depicted in squares connected by arrows to the block where this call happened. For example, the $submitVote$ method from the orange chain is called twice, and later $transferVotes$ method is called. These method calls are also through the IOTA-ROS2 bridge.

\subsection{Partition-Tolerant Collaborative Decision Making}

As we have described earlier, current solutions that leverage DLTs in robotics are not able to deal with network partitions. Indeed, if two robots are meant to share data regularly on a blockchain, but they remain disconnected for part of the data gathering process, a problem raises in terms of how can the individual chains be merged. We discuss here how asynchronous calls in the ISCP can solve this issue. Our objective is to enable multiple robots to achieve consensus in terms of global mission parameters or variables (e.g., a shared map or a bijective role allocation), while maintaining local consensus within their internal systems or local subnetwork.

\begin{lstlisting}[language=Solidity,backgroundcolor = \color{white},
caption={General smart contract definition (Ethereum's Solidity) for a collaborative decision making implementation.},
label={lst:gen_sc}]
pragma solidity ^0.8.7;
    
contract MainContract {
        
    VOTE[] votes;
    	
    function submitVote (VOTE memory vote) {
        // Some code goes here
        votes.push(vote);
        // Rest of the code goes here
    }
    	
    function aggregate () returns result {
        // Implementation goes here
        return result;
    }
}
\end{lstlisting}

Collaborative decision-making processes can be typically implemented on a smart contract following the example in Listing.~\ref{lst:gen_sc}. This example, for the Ethereum blockchain, has been followed by different works in the liteature. The collaborative decision-making process then proceeds as follows. Every entity that is participating in the task has a vote. These votes could be any type of data based on the goal they are going to achieve. Every participant can submit their vote with the $submitVote$ function of the smart contract. $aggregate$ is a view function (only reads data from the ledger and doesn't change anything) used for aggregating the votes to get the final result of consensus.

Such collaborative decision making processes which can be formulated as a smart contract similar to the general form mentioned above can become partition tolerant. Partition tolerance is achieved by the multiple chains introduced by ISCP. In order to make a smart contract partition tolerance, we propose that it should be divided into two separate smart contracts. First, a \textsc{Controller Swarm} smart contract should be deployed on the \textsc{Controller} chain. Second, every group of robots that may at some point be disconnected from the core network for a certain amount of time should create their own chain and deploy a new instance of an \textsc{Agent Swarm} smart contract on the chain.

The \textsc{Agent Swarm} contract is responsible to store the votes on the ledger, so that they are saved in case that these robots disconnect from the rest of the network through the $submitVote$ function. In addition, this smart contract provides the $transferVotes$ function that can be called by any member on this chain to transfer the stored votes to the \textsc{Controller} chain. This function uses an asynchronous call to the $submitVotes$ function of \textsc{Controller Swarm} smart contract. The \textsc{Controller Swarm} contract implements the $aggregator$ function that aggregates all the votes from the different chains. These smart contracts are illustrated in Listing.~\ref{lst:controllerswarm_sc} and Listing.~\ref{lst:agentswarm_sc}.

\begin{lstlisting}[language=bash,backgroundcolor = \color{white},
caption={ControllerSwarm smart contract schema.},
label={lst:controllerswarm_sc}]
name: ControllerSwarm
structs:
  VOTE:
    # Data structure of votes goes here
state:
  votes: VOTE[]
  # Other state variables goes here
funcs:
  submitVotes:
    params:
      votes: VOTE[]
  aggregate:
    results:
      result: RESULT
  # Other functions may be also defined here
\end{lstlisting}

\begin{lstlisting}[language=bash,backgroundcolor = \color{white},
caption={\textsc{Agent Swarm} smart contract schema.},
label={lst:agentswarm_sc}]
name: AgentSwarm
structs:
  VOTE:
    # Data structure of votes goes here
state:
  votes: VOTE[]
  # Other state variables goes here
funcs:
  submitVote:
    params:
      vote: VOTE
  transferVotes: {}
  # Other functions may also be defined here
\end{lstlisting}

\subsection{Eventual Consistency}

Based on Brewer's CAP conjecture for a distributed system~\cite{brewer2000towards}, we cannot achieve consistency, availability, and partition tolerance simultaneously. As It is proposed to make the collaborative decision making tasks partition tolerant in a distributed manner, the consistency will not be acquired anymore. As mentioned in the previous section, the $submitVotes$ function of the \textsc{Agent Swarm} smart contract is called when the chain members are again connected to the rest of the network. Therefore, in this case, the system has eventual consistency. Every group of swarms on a chain can be disconnected for a while, but when they are connected to the \textsc{Controller} chain, the system will reach an eventual consistency.
This assumption does not lead to any degradation of the functionality of the robotic system. Indeed, if a robot or group of robots remains disconnected from the \textsc{Controller} chain, then they can maintain operation within the ledgers defined in their own network partition, unrelated to the rest of the system. If the disconnection is a result of a malfunction, then it is out of the scope of this work to provide connectivity maintenance or recovery methods, with the robotics literature containing solutions to such challenges.

To achieve more robust multi-robot systems using DLTs, it is worthy to prefer partition tolerance over consistency. However it is possible to have eventual consistency that is sufficient in many scenarios. If we want to choose consistency, then we might loose some part of the data. In next section, we will demonstrate a scenario that loosing data how can affect the overall decision making problem.

\begin{figure*}[t]
    \centering
    \includegraphics[width=\textwidth]{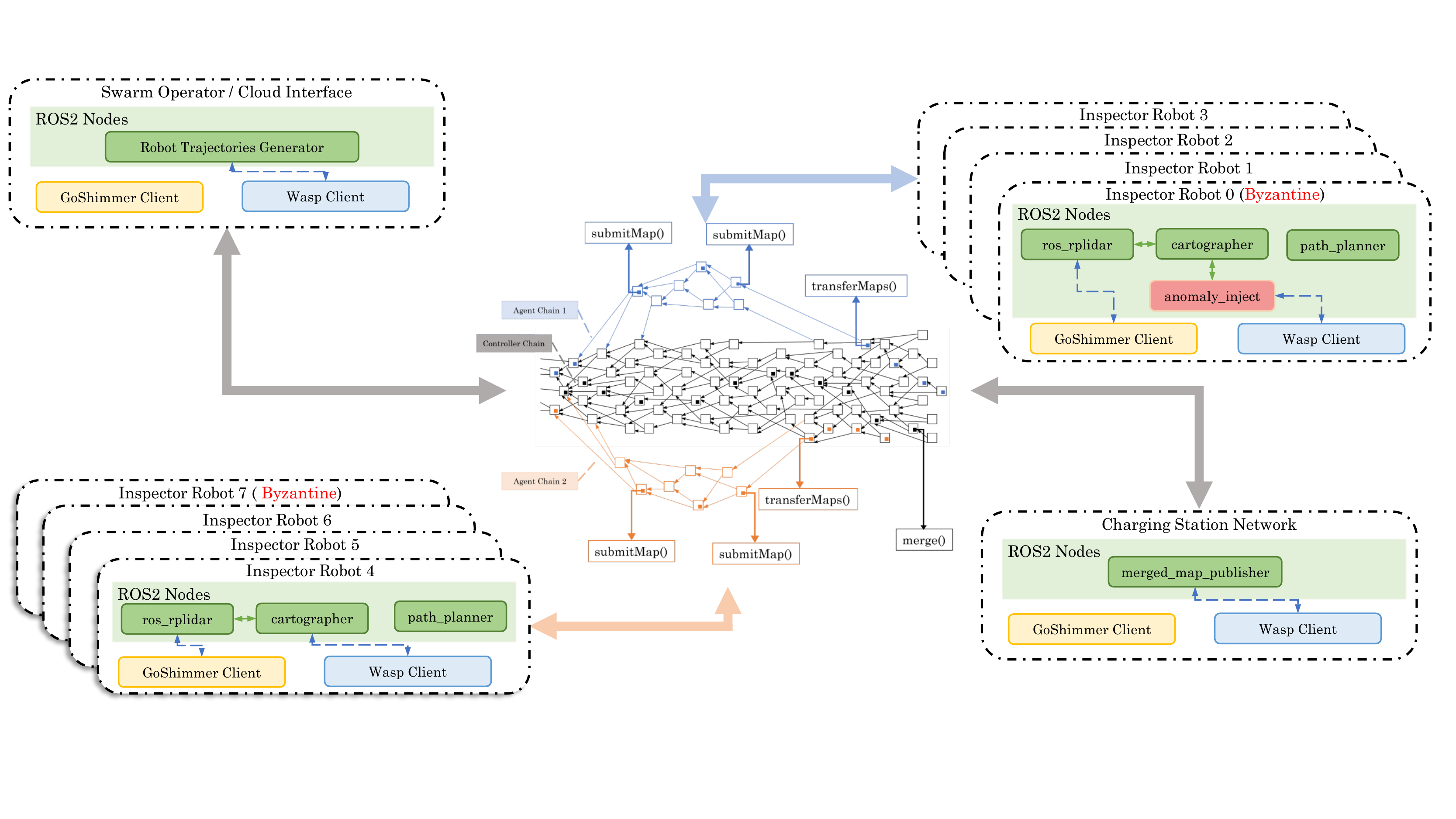} \\[-3em]
    \caption{Integration of ROS\,2 and IOTA smart contract for the cooperative mapping application use case.}
    \label{fig:arc_exp}
\end{figure*}

\section{Experimental Results}\label{sec:results}

In this section, we focus on a specific proof of concept to demonstrate the usability and applicability of the proposed methods. The vast majority of indoor mobile robots today use 2D lidars for navigation and mapping, with mapping being a key step for a robot to achieve situational awareness. 
We choose cooperative mapping as a representative task involving consensus (agreement on the final map of the operational environment) and the aggregation of data from different robots (the individual maps). This task includes role allocation, and agreement between the entities. At the same time, the individual maps can be checked against the real-time sensor data published in the Tangle. Finally, collaborative mapping is an excellent example of a consensus problem requiring a partition-toleran implementation for real-world applications. The network partitioning is highly probable due to undiscovered heterogeneous environments. Also, byzantine entities might alter the collaboratively created map to exclude some areas to be navigated, or even expose other robots to real hazards by altering their local maps. In summary, we choose a simple and intuitive approach for collaborative mapping to demonstrate the method proposed in the previous section.

\begin{algorithm}

    \caption{Compare two maps}\label{alg:compare_maps}
    \KwIn{$M_{l \times w}, M'_{l \times w}$}
    \KwOut{$True$ or $False$}
    $comply \gets True$\;
    $T \gets 0.2$\;
    $T' \gets 0.2$\;
    \ForEach{$W_{n \times n}$ \textbf{in} $M_{l \times w}$}{
        \ForEach{$W'_{n \times n}$ \textbf{in} $M'_{l \times w}$}{
            $\mathbb{S} \gets \{w_{i \times j} = w'_{i \times j}, w_{i \times j} \neq U\} $\;
            $\mathbb{D} \gets \{w_{i \times j} \neq w'_{i \times j}, w_{i \times j} \neq U, w'_{i \times j} \neq U\} $\;
            $\mathbb{U} \gets \{w_{i \times j} = U\} $\;
            $\mathbb{U'} \gets \{w'_{i \times j} = U\} $\;
            \If{$\dfrac{|\mathbb{U}|}{n^2} < T \land  \dfrac{|\mathbb{U'}|}{n^2} < T  \land \dfrac{|\mathbb{D}|}{|\mathbb{D}| + |\mathbb{S}|} > T'$}{
                $comply \gets False$
            }
        }
    }
    \Return{comply}
    
\end{algorithm}

\begin{algorithm}

    \caption{Smart contract map submission}\label{alg:submit_map}
    \KwIn{$M_{l \times w}$}
    \KwData{$\mathbb{M}$ List of submitted maps}
    
    $complies \gets 0$\;
    $contracts \gets 0$\;
    \ForEach{$M'_{l \times w}$ \textbf{in} $\mathbb{M}$}{
        $comply \gets $ \textit{compare} $(M_{l \times w},M'_{l \times w})$\;
        \If{comply}{
            $complies \gets complies + 1$\;
            \textit{complied}$(M'_{l \times w}) \gets$ \textit{complied}$(M'_{l \times w}) + 1$\;
        }
        \Else{
            $contracts \gets contracts + 1$\;
            \textit{contracted}$(M'_{l \times w}) \gets$ \textit{contracted}$(M'_{l \times w}) + 1$\;
        }
    }
    \textit{complied}$(M_{l \times w}) \gets complies$\;

    \textit{contracted}$(M_{l \times w}) \gets contracts$\;
    $\mathbb{M}.$\textit{append}$(M_{l \times w})$\;
    
\end{algorithm}

\begin{algorithm}

    \caption{Smart contract merge function}\label{alg:merge}
    \KwData{$\mathbb{M}$ List of submitted maps}
    
    $global\_map \gets \emptyset$\;
    \ForEach{$M_{l \times w}$ \textbf{in} $\mathbb{M}$}{
    
        \If{\textit{complied}$(M'_{l \times w}) <$ \textit{contracted}$(M_{l \times w})$}{
            \textit{agent}$(M_{l \times w}) \gets byzantine$\;
        }
        
        \If{\textit{agent}$(M_{l \times w})$ \textbf{is not} $byzantine$}{
            $global\_map.$\textit{append}$(M_{l \times w})$\;
        }
    }
    \Return{$global\_map$}\;

\end{algorithm}

\subsection{Partition-Tolerant Collaborative Mapping}

For the collaborative mapping application, we consider a rectangular area to be mapped. We assume that the rectangle has a known size of $L\times W$. We also assume that there are multiple robots that are going to participate in this task as a service. The area is divided into $l\times w$ sized cells. The cells are considered to be small enough so that the topology does not change significantly from different viewpoints, and so that a single lidar scan is enough for mapping the cell. Each cell should be visited by $k\geq3f+1$ robots, such that at most $f$ out of them are byzantine. 

First, robots should register their identity to take part in the mapping task. Second, each robot is assigned to a randomly selected cell in the area. Third, the robot travels to the assigned cell to map it. Fourth, the robot submits the built map to the smart contract. The smart contract based on the received map merges them.

We define every local map as a 2D matrix of size $M_{l\times w}$. The resolution for creating the map is a fixed and predefined value. Every entry of the matrix is filled by the robot with out of three values ($\{O, F, U\}$) representing occupied, free, and unknown, respectively. 

To implement this partition-tolerant collaborative mapping on a smart contract, we define five main functions. First, a role allocation function to assign a random cell to each robot which is going to participate in the inspection. This function is responsible to assign each cell for at least $k\geq3f+1$ robots to tolerate $f$ byzantine robot. Second, a function that is responsible of comparing two maps as defined in Algorithm~\ref{alg:compare_maps}. This function will return a true or false value indicating where the two input maps are complying (true) or conflicting (false). This comparison function will then be used in the submission function. The third function is the submission function, defined in Algorithm~\ref{alg:submit_map}. For every new map that is submitted to the smart contract, it is compared to the all previous submitted ones. For each map, two values are stored. The $comply()$ value of the map, representing the number of maps complying with this map, and the $conflict()$, defined in the same way for the number of maps with which it conflits.

The fifth function, namely map merging, is defined in Algorithm~\ref{alg:merge}. The merging function is responsible for merging the maps. In this context, merging is the $aggregate$ function of the general smart contract defined in the previous section. Equivalently, the generic $votes$ defined earlier are now maps in this application, and the generic $submitVote$ function is implemented through the map submission function.

\begin{figure*}[t]
    \centering
    \includegraphics[width=\textwidth]{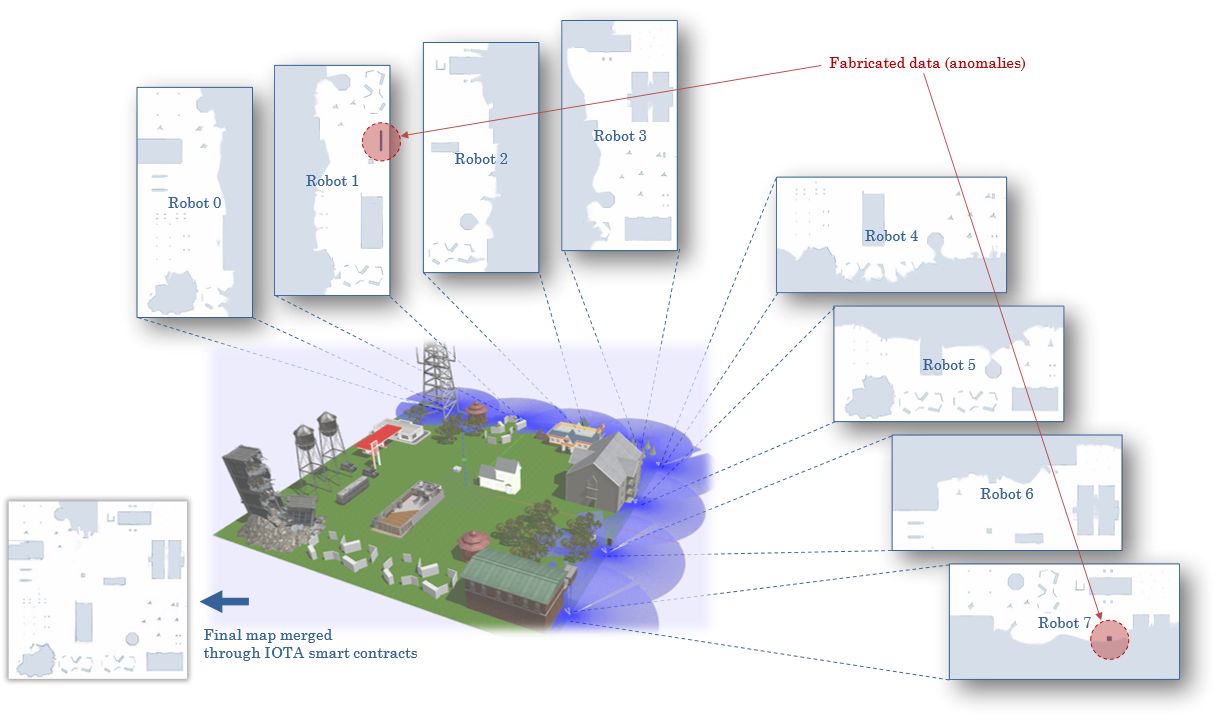}
    \caption{Mapping results in the Gazebo simulation where eight robots map a small town. The dark blue lines represent the lidar scan range for each of the robots; the robot paths are omitted to improve visualization. Each individual map is generated by the robots locally using cartographer. The final map obtained through the IOTA smart contracts is shown in the bottom left, where the anomalies detected in the data submitted by two of the robots have been effectively eliminated.}
    \label{fig:screenshot}
\end{figure*}

\begin{figure*}[t]
    \centering
    \includegraphics[width=\textwidth]{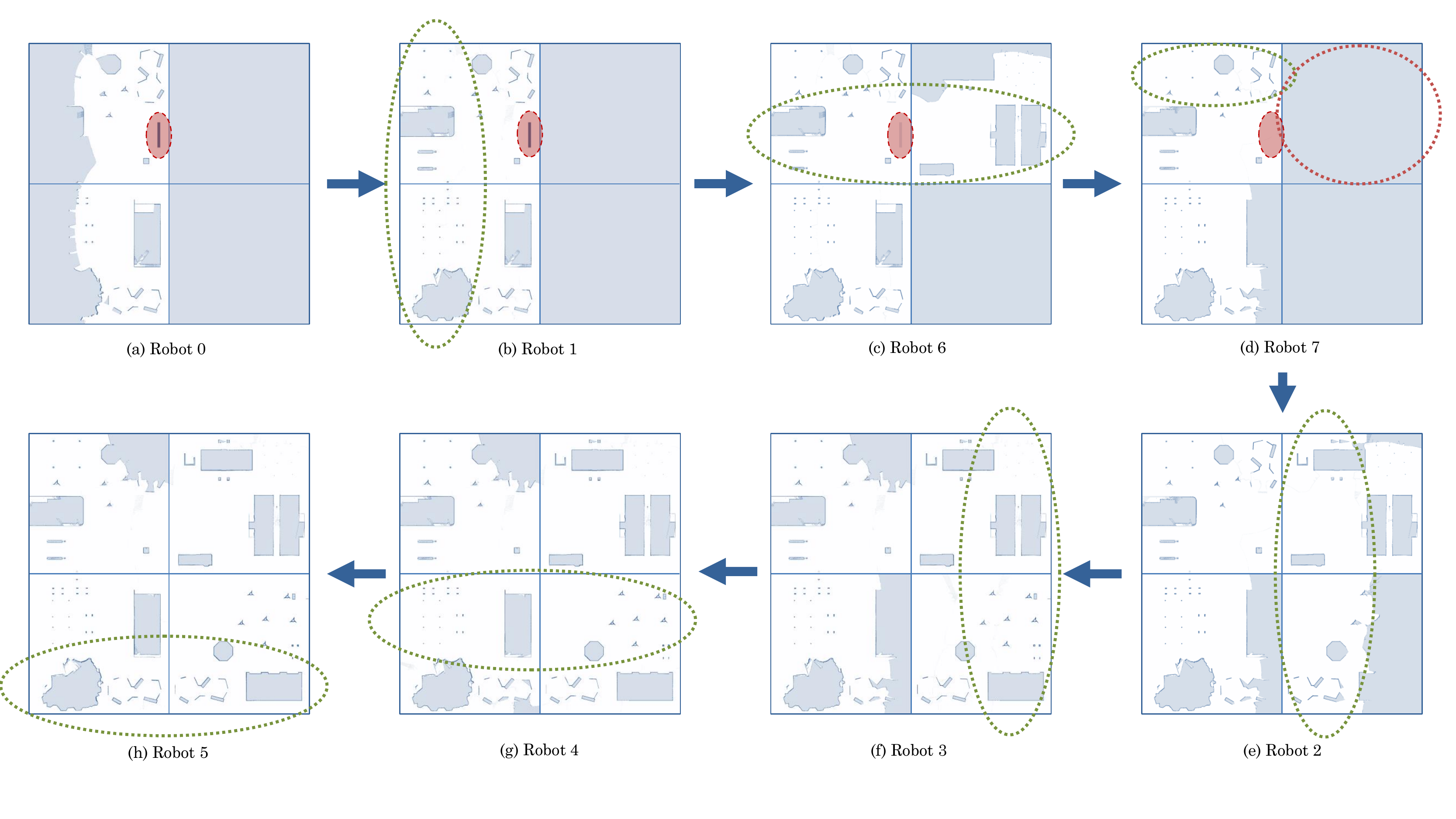}
    \caption{Illustration of the map merging process as different robots submit their local maps in time. Once \textit{Robot\,0} and \textit{Robot\,1} have submitted their maps to the smart contract, the overall map is updated as these two maps are not conflicting. The process continues with other local maps, with anomalies being effectively eliminated once enough data is available about the corresponding map cells. An alternative implementation might wait for a certain number of robots to submit their maps before merging the data, without changing the core processes.}
    \label{fig:sc_maps}
\end{figure*}

\subsection{Multi-Robot Gazebo Simulation}

For the purpose of testing the proposed partition-tolerant collaborative mapping task with a larger number of robots, a small town-like environment is designed in the Gazebo simulator. The simulation environment is shown in Fig.~\ref{fig:screenshot}. Eight robots are deployed in the simulator to map the town. We set two of these robots as byzantine agents. Their objective is to try to \textit{fool} other robots by inserting certain random walls on their local town maps. Each robot scans the surrounding environment with a 2D lidar while navigating the town on a predefined path. The output of the 2D lidar and the odometry data generated by Gazebo simulator are used for mapping using the Cartographer ROS\,2 implementation~\cite{hess2016real}. A ROS\,2 node is implemented in order to emulate the behavior of byzantine agents. The final local maps (either directly the output of the Cartographer node or the altered maps generated by the nodes running in the byzantine agents) are submitted by robots to the Controller Swarm smart contract. The maps that are submitted by each robot are shown in Fig.~\ref{fig:screenshot}, each covering a different area of the town and with the lack of a global view. After inspecting the assigned cell, the robot will submit its map to the smart contract. Another ROS\,2 node is implemented to call the merge function of the smart contract and publish the results on a ROS\,2 topic. The merged map in different stages is illustrated in Fig.~\ref{fig:sc_maps}. In this experiment \textit{Robot\,0} and \textit{Robot\,7} are byzantine robots. The maps submitted by these robots are eliminated in the overall merged map when enough data is available about the sections where byzantine agents alter the real maps. It is worth noting at this point that inserted data may temporarily appear in the merged map until a high enough number of robots submit a map that conflicts with the map submitted by the byzantine robot. For a specific application and collaborative decision-making process implementation, a byzantine behaviour can be potentially designed in an adversarial manner to surpass the detection mechanisms in the smart contract. However, our focus here is not on defining a robust byzantine agent detection strategy bus instead on introducing a general way of building trust with IOTA through a partition-tolerant implementation. Therefore, more advanced smart contracts specifically designed to detect altered data are out of the scope of this work.

The simulation results show that our method effectively enables both byzantine-tolerant and partition-tolerant collaborative mapping under certain conditions. Through the simulation, we effectively emulate network disconnections as robots only submit their local maps once their path is finalized. In this implementation, each robot individually forms its own Wasp committee. In practice, several robots operating in nearby cells can form a common Wasp committee.


\begin{figure}[t]
    \centering
    \setlength{\figureheight}{0.35\textwidth}
    \setlength{\figurewidth}{.48\textwidth} 
    \scriptsize{
\begin{tikzpicture}

\definecolor{darkgray176}{RGB}{176,176,176}
\definecolor{orange}{RGB}{255,165,0}
\definecolor{yellow}{RGB}{255,255,0}
\definecolor{midnight_blue}{HTML}{003B73}
\definecolor{blue_gray}{HTML}{0074B7}
\definecolor{dark_blue}{HTML}{60A3D9}
\definecolor{baby_blue}{HTML}{C3E0E5}

\begin{axis}[ 
    width = \figurewidth,
    height = \figureheight,
    tick align=outside,
    tick pos=left,
    axis line style={white},
    legend style={fill opacity=0.8, draw opacity=1, text opacity=1, 
    draw=white!80!black},
    tick align=outside,
    tick pos=left,
    x grid style={darkgray176},
    xtick style={color=black},
    y grid style={darkgray176},
    ymajorgrids,
    ymin=6.057214419123e-05, ymax=0.077709654801712,
    ytick style={color=black},
    xmin=-0.735, xmax=7.735,
    xtick={0,1,2,3,4,5,6,7},
    xticklabels={$R_0$,$R_1$,$R_6$,$R_7$,$R_2$,$R_3$,$R_4$,$R_5$},
    ylabel={Execution Time (seconds)},
    ymin=0, ymax=2.19728293418884,
    ytick style={color=black}  
]
\draw[draw=none,fill=midnight_blue] (axis cs:-0.35,0) rectangle (axis cs:0,0.02);
\draw[draw=none,fill=midnight_blue] (axis cs:0.65,0) rectangle (axis cs:1,0.688796043395996);
\draw[draw=none,fill=midnight_blue] (axis cs:1.65,0) rectangle (axis cs:2,0.890658855438232);
\draw[draw=none,fill=midnight_blue] (axis cs:2.65,0) rectangle (axis cs:3,1.68359923362732);
\draw[draw=none,fill=blue_gray] (axis cs:3.65,0) rectangle (axis cs:4,0.02);
\draw[draw=none,fill=blue_gray] (axis cs:4.65,0) rectangle (axis cs:5,0.701498508453369);
\draw[draw=none,fill=dark_blue] (axis cs:5.65,0) rectangle (axis cs:6,1.34033226966858);
\draw[draw=none,fill=dark_blue] (axis cs:6.65,0) rectangle (axis cs:7,2.09265041351318);
\draw[draw=none,fill=dark_blue] (axis cs:2.77555756156289e-17,0) rectangle (axis cs:0.35,0.02);
\draw[draw=none,fill=dark_blue] (axis cs:1,0) rectangle (axis cs:1.35,0.706751585006714);
\draw[draw=none,fill=baby_blue] (axis cs:2,0) rectangle (axis cs:2.35,0.03);
\draw[draw=none,fill=baby_blue] (axis cs:3,0) rectangle (axis cs:3.35,0.247498035430908);
\draw[draw=none,fill=baby_blue] (axis cs:4,0) rectangle (axis cs:4.35,0.910317182540894);
\draw[draw=none,fill=baby_blue] (axis cs:5,0) rectangle (axis cs:5.35,1.5749294757843);
\draw[draw=none,fill=blue_gray] (axis cs:6,0) rectangle (axis cs:6.35,1.39223670959473);
\draw[draw=none,fill=blue_gray] (axis cs:7,0) rectangle (axis cs:7.35,2.09265041351318);
\end{axis}

\end{tikzpicture}}
    \caption{Execution time of $SubmitVotes$ method of \textsc{Controller Swarm} smart contract for each robot. The simulation environment is divided in four cells or quadrants. Each robot navigates mainly through two of the quadrants of the map and submits two separate maps. The four bar colors in the graph represent these four cells. Every map submitted is compared on a cell-by-cell basis only when there is an intersection between maps.}
    \label{fig:exec_time}
    \vspace{-0.42em}
\end{figure}
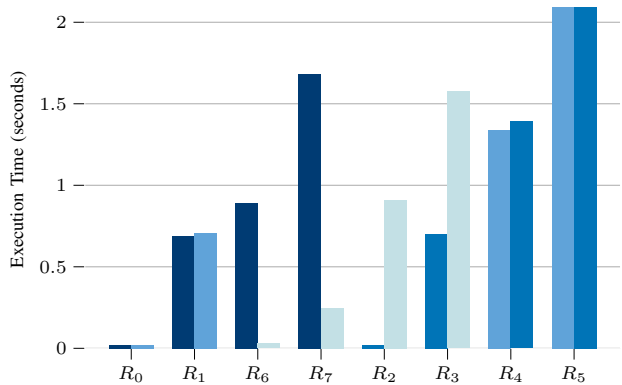

\begin{figure}[t]
    \centering
    \includegraphics[width=0.45\textwidth]{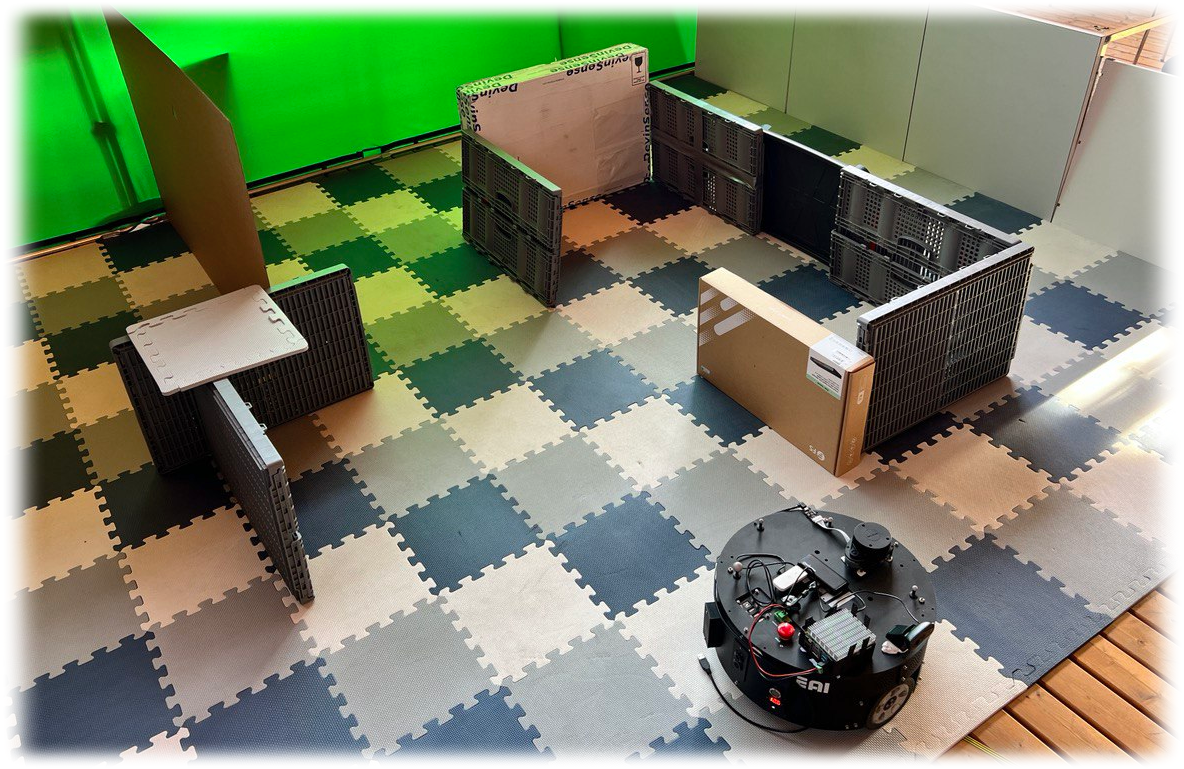}
    \caption{Portion of the maze created for the mapping experiments. The Dashgo UGV is also seen in the image.}
    \label{fig:maze}
    \vspace{-0.42em}
\end{figure}

\begin{figure*}[t]
    \centering
    \includegraphics[width=0.99\linewidth]{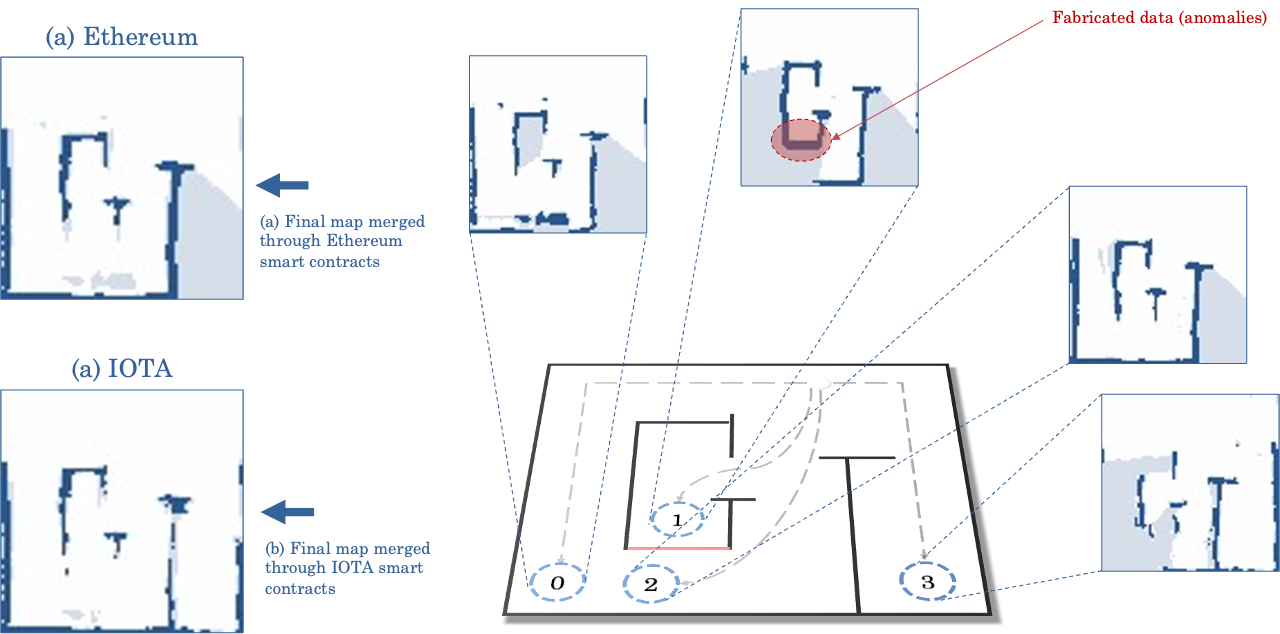}
    \caption{Real-world maps merged by IOTA and Ethereum smart contracts.}
    \label{fig:real_world_exp}
\end{figure*}

\subsection{Real-World Experiment}

For real-world experiments we used a Dashgo indoor UGV with an RPLIDAR A1 installed on top of it. For localization, we rely on an Optitrack MOCAP system providing global, real-time, 6D pose estimation through ROS\,2 topics. We create a small maze to be navigated by robots. For the sake of simplicity, we map different areas of the maze iteratively, adding or removing physical elements in between to emulate altered sensor data. The maze and the Dashgo UGV assembly can be seen in Fig.~\ref{fig:maze}. In practice, the same robot is operated four times to create the data for four different virtual agents. This is done without loss of generality from the perspective of the methods introduced in this paper. Four different GoShimmer and Wasp nodes are deployed in two sub-networks to demonstrate the partition tolerance of the methods, with these sub-networks  simulating network disconnections. These nodes are deployed on an UP Squared board placed on top of the Dashgo UGV. Also, an UP Xtreme board is used to play the role of charging station with its own GoShimmer and Wasp nodes. The hardware specifications of these boards are listed in Table.~\ref{tab:hardware}. 
In this experiment, \textit{Robot\,3} will simulate the network disconnection. GoShimmer and Wasp nodes associated to \textit{Robot\,3} are deployed on a Docker network. The \textit{Robot\,3} Wasp node will create a chain and deploy the \textsc{Agent Swarm} smart contract. Other GoShimmer and Wasp nodes will use another Docker network with a chain which another instance of \textsc{Agent Swarm} smart contract is deployed on. GoShimmer and Wasp nodes of the charging station also has their own Docker network and \textsc{Controller Swarm} smart contract deployed on a chain. These Docker networks communicate through an Docker overlay network. The White box in Fig.~\ref{fig:maze} is inserted as an anomaly while \textit{Robot\,1} is navigating, which corresponds to the byzantine robot mapping the area.

The raw lidar scan is published by GoShimmer nodes on the Tangle. \textit{ROS\,2 LaserScan} messages, which are approximately 1.5\,KB each, are published with a 5.5\,Hz rate. The CPU and memory consumption of every board is reported in Table.~\ref{tab:perf}. The map created by each robot is illustrated in Fig.~\ref{fig:real_world_exp}.

\begin{table}[t]
    \centering
    \caption{Hardware Specifications of UP Squared and Xtreme boards used for real-world experiments.}
    \label{tab:hardware}
    \renewcommand{\arraystretch}{1.23}
    \small
    \begin{tabular}{@{}lccl@{}}
        \toprule
         & \textbf{UP Squared} & \textbf{Up Xtreme} \\
        \midrule
        \textbf{CPU} & Intel Atom x7-E3950 & Intel Core i7-8665UE \\
        \textbf{Mem / Disk} & 8 GB / 64 GB & 16 GB / 64 GB \\
        \bottomrule
    \end{tabular}
\end{table}

\begin{table}[t]
    \centering
    \caption{CPU and memory consumption of GoShimmer nodes on UP Squared and Xtreme boards}
    \label{tab:perf}
    \renewcommand{\arraystretch}{1.23}
    \small
    \begin{tabular}{@{}lccl@{}}
        \toprule
         & \textbf{UP Squared} & \textbf{Up Xtreme} \\
        \midrule
        \textbf{GoShimmer avg. CPU} & 25.85\% & 20.16\% \\
        \textbf{GoShimmer avg. MEM)} \hspace{1.42em} & 213.65 MB & 228.8 MB \\
        \bottomrule
    \end{tabular}
\end{table}

To see the difference between IOTA and Ethereum and how partition-tolerance can benefit this mapping process, the maps are submitted to smart contracts deployed on both Ethereum and IOTA networks. 
The map merging result of the Ethereum and IOTA smart contracts are illustrated in Fig.~\ref{fig:real_world_exp}. In the merged map generated through the Ethereum smart contract we can observe that the map submitted by the robot which was in different sub-network was eliminated. This is effectively caused because only one of the two chains created when robots are disconnected remains at the end of the mission (Ethereum discards the shortest chain, also defined as the chain with the lowest accumulated computational complexity). With the IOTA-based implementation, this part of the map is included and the anomaly wall effectively removed at the same time.

These results demonstrate the following. First, the IOTA-based implementation is superior to more traditional Ethereum implementation in terms of supporting network partitions. In practice, this means that whenever robots disconnect in the real world, their data is not necessarily lost. We quantify this with the percentage of mapped area that results from IOTA and Ethereum smart contracts and listed in Table~\ref{tab:map_percentages}.
IOTA smart contracts achieve higher percentage of mapped area than Ethereum smart contracts and even the union of robot maps by masking the anomalies.

With the vast majority of the literature in DLT integrations for robotics relying on traditional blockchains, this work solves one of the key practical problems stopping more widespread use of this technology. Second, the IOTA-based implementation is effectively able to detect and neutralize the behaviour from the byzantine agents. This is a novel results from the point of view of the technology in use, albeit several works in the literature showcase such ability in Ethereum-based solutions. Nonetheless, this result also shows that the IOTA-based methods maintain the functionality of existing research while extending applicability. Finally, a third conclusion from the reported results is that this work opens the door to more scalable and wider use of DLTs within distributed robotic systems. While this is not proved in this paper, the DAG-based architecture of IOTA and the nature of its solutions with respect to Ethereum and other traditional blockchains mean that more scalable solutions for more resource-constrained real-world devices can be designed and developed. It is worth noting that the Ethereum community is also working, with different methods, towards a more scalable DLT, and proof-of-authority implementations already bring some benefits, albeit not the network partition tolerance we look for in this work. 

\begin{figure}[t]
    \centering
    \begin{subfigure}[t]{0.30\textwidth}
        \centering
        \setlength{\figureheight}{1.2\textwidth}
        \setlength{\figurewidth}{1\textwidth} 
        \scriptsize{\input{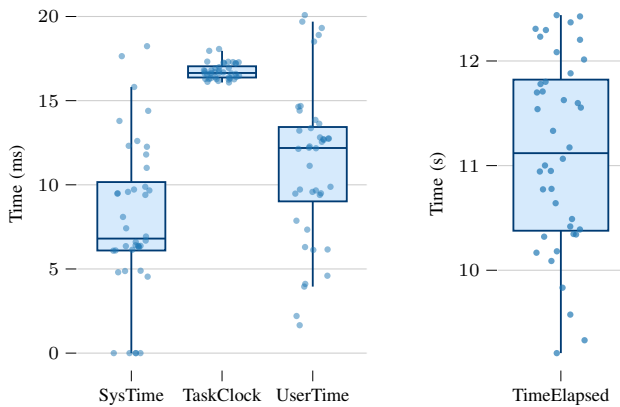}}
        \caption{CPU utilization time of the $SubmitVotes$ method measured with the Linux perf tool.}
        \label{fig:sc_exec_time}
    \end{subfigure}
    \hfill
    \begin{subfigure}[t]{0.18\textwidth}
        \centering
        \setlength{\figureheight}{2\textwidth}
        \setlength{\figurewidth}{1\textwidth} 
        \scriptsize{
\begin{tikzpicture}

\definecolor{darkslategray51}{RGB}{51,51,51}
\definecolor{lavender235}{RGB}{205,205,205}
\definecolor{midnight_blue}{HTML}{003B73}
\definecolor{blue_gray}{HTML}{0074B7}
\definecolor{dark_blue}{HTML}{60A3D9}
\definecolor{baby_blue}{HTML}{C3E0E5}
\definecolor{light_blue}{RGB}{214,234,252}

\begin{axis}[
axis line style={white},
width=\figurewidth,
height=\figureheight,
tick align=outside,
tick pos=left,
x grid style={white},
xmin=0.5, xmax=1.5,
xtick style={color=black},
xtick={1},
xticklabels={TimeElapsed},
ylabel=\textcolor{black}{Time (s)},
ymajorgrids,
y grid style={lavender235},
ymin=9.0462600299175, ymax=12.5999393166706,
yminorgrids,
ytick style={color=black}
]
\path [draw=midnight_blue, line width=0.7pt]
(axis cs:1,11.82179796)
--(axis cs:1,12.43840844);

\path [draw=midnight_blue, line width=0.7pt]
(axis cs:1,10.37772257)
--(axis cs:1,9.208053591);

\path [draw=midnight_blue, fill=light_blue, line width=0.7pt]
(axis cs:0.625,11.82179796)
--(axis cs:0.625,11.82179796)
--(axis cs:1.375,11.82179796)
--(axis cs:1.375,10.37772257)
--(axis cs:0.625,10.37772257)
--cycle;

\path [draw=midnight_blue, line width=0.708981540362206pt]
(axis cs:0.625,11.11895322)
--(axis cs:1.375,11.11895322);

\addplot [
  mark=*,
  only marks,
   fill opacity=0.7,
  draw opacity=0,
  mark size=1.2pt,
  scatter,
  scatter/@post marker code/.code={%
  \endscope
},
  scatter/@pre marker code/.append style={/tikz/mark size=\perpointmarksize},
  scatter/@pre marker code/.code={%
  \expanded{%
  \noexpand\definecolor{thispointdrawcolor}{RGB}{\drawcolor}%
  \noexpand\definecolor{thispointfillcolor}{RGB}{\fillcolor}%
  }%
  \scope[draw=thispointdrawcolor, fill=thispointfillcolor]%
},
  visualization depends on={\thisrow{sizedata} \as\perpointmarksize},
  visualization depends on={value \thisrow{draw} \as \drawcolor},
  visualization depends on={value \thisrow{fill} \as \fillcolor}
]
table{%
x  y  draw  fill  sizedata
0.96680880188103 12.0843793931414 31,119,180 31,119,180 2.5
1.08812979737686 10.4896346061511 31,119,180 31,119,180 2.5
0.800045749926938 12.3073235198124 31,119,180 31,119,180 2.5
0.920933029052736 10.9490290879831 31,119,180 31,119,180 2.5
0.858702356326845 10.772911505572 31,119,180 31,119,180 2.5
0.836935437907519 12.23204836452 31,119,180 31,119,180 2.5
0.874504084551068 11.0013920371654 31,119,180 31,119,180 2.5
0.938224290817219 11.3321862351049 31,119,180 31,119,180 2.5
0.958706989692268 10.6407386892276 31,119,180 31,119,180 2.5
1.01552669360134 11.0651530763705 31,119,180 31,119,180 2.5
0.967677805761318 9.20779090658809 31,119,180 31,119,180 2.5
1.0740878001587 10.4182099559972 31,119,180 31,119,180 2.5
0.881780899892607 12.2958107462803 31,119,180 31,119,180 2.5
1.15124697455638 12.425780103237 31,119,180 31,119,180 2.5
0.810955037279171 11.6989695602453 31,119,180 31,119,180 2.5
1.06818700407136 11.1734337098733 31,119,180 31,119,180 2.5
0.966921920946851 12.4381396124669 31,119,180 31,119,180 2.5
1.0234759313783 11.6261344003029 31,119,180 31,119,180 2.5
0.856154775438093 11.7083488012293 31,119,180 31,119,180 2.5
0.879240595633952 11.8009278110752 31,119,180 31,119,180 2.5
1.12029782747021 10.3419253776617 31,119,180 31,119,180 2.5
1.18730463028776 9.33081630296862 31,119,180 31,119,180 2.5
0.925369671263697 10.0884284948405 31,119,180 31,119,180 2.5
1.07692904626773 12.3693108369748 31,119,180 31,119,180 2.5
1.15055566091842 10.3896547748896 31,119,180 31,119,180 2.5
1.15784266540154 11.5545121468123 31,119,180 31,119,180 2.5
0.834017684547911 10.9435542317634 31,119,180 31,119,180 2.5
0.815621913293153 11.5397497882266 31,119,180 31,119,180 2.5
0.867932167825828 10.3206560868039 31,119,180 31,119,180 2.5
1.15125700137177 12.2027081547256 31,119,180 31,119,180 2.5
0.83933873353322 11.7797740562477 31,119,180 31,119,180 2.5
0.968443050002021 10.180909464774 31,119,180 31,119,180 2.5
1.1831558120602 12.014051078274 31,119,180 31,119,180 2.5
1.01326611398921 9.83311634250436 31,119,180 31,119,180 2.5
1.07675084558019 11.8820879087271 31,119,180 31,119,180 2.5
0.926206252402425 10.7775236437856 31,119,180 31,119,180 2.5
1.07460037107263 9.57604713469173 31,119,180 31,119,180 2.5
1.13385026875895 11.5975032791652 31,119,180 31,119,180 2.5
0.807315310937677 10.1676017999709 31,119,180 31,119,180 2.5
1.10005772597799 10.3464098551425 31,119,180 31,119,180 2.5
};
\end{axis}

\end{tikzpicture}}
        \caption{Confirmation time for the $SubmitVotes$ method.}
        \label{fig:sc_confirmation_time}
    \end{subfigure}
    \caption{Distribution of the execution and confirmation time for the $SubmitVotes$ method of the \textsc{Controller Swarm} smart contract for each robot over multiple experimental runs. The actual CPU utilization time is in the range of a few \textit{ms}, while the default IOTA network configuration results in confirmation times slightly over 10\,s.}
    \label{fig:execution_times}
    \vspace{-1em}
\end{figure}

\begin{table}[t]
    \centering
    \caption{Percentage of total area mapped by each individual robot, and percentage of total area generated by the Ethereum and IOTA smart contracts merging the individual maps.}
    \label{tab:map_percentages}
    \renewcommand{\arraystretch}{1.23}
    \small
    \begin{tabular}{@{}ccl@{}}
        \toprule
        \textbf{Map generator} & \textbf{Mapped area} (\%) & \textbf{Note} \\
        \midrule
        \textit{Robot\,0} & 77.78 \% & \\
        \textit{Robot\,1} & 65.55 \% & \footnotesize{\textit{Byzantine agent}} \\
        \textit{Robot\,2} & 83.95 \% & \\
        \textit{Robot\,3} & 76.39 \% & \footnotesize{\textit{Disconnected agent}} \\
        \midrule
        $\cup_{i=0}^{3}Robot_i$ & 94.64 \% \\
        Ethereum SC & 82.2. \% & \footnotesize{\textit{Agents 0 + 2}} \\
        IOTA SC & 95.50 \% & \footnotesize{\textit{Agents 0 + 2 + 3}} \\
        \bottomrule
    \end{tabular}
\end{table}


\section{Discussion}
\label{sec:discussion}

In addition to the results above, we measure the CPU utilization time and the confirmation time for one of the smart contract methods. Figure~\ref{fig:execution_times} shows these results. The CPU utilization time indicates the actual impact of the IOTA smart contract in terms of resource utilization. Using the Linux perf tool, we measure the execution time of the \textit{SubmitVotes} method of the \textsc{Controller Swarm} smart contract. Figure~\ref{fig:sc_exec_time} shows that the typical execution time is in the range of 10\,ms to 20\,ms, revealing the potential for scalability and the ability of running multiple smart contracts in parallel. The use of DLTs and the decentralized consensus mechanisms in them, however, introduces a certain delay until a transaction is \textit{confirmed}. In practice, this means that the transaction, in this case the result of a smart contract method, is confirmed by the nodes in the Wasp Committee. Even if the actual network delay is small in the experiments, the default configuration of the IOTA research network introduces two delays of about 5\,s each to avoid double spending. This results in confirmation times over 10\,s reported in Fig.~\ref{fig:sc_confirmation_time}. If a faster solution is required, the network configuration can be changed to reduce the confirmation time, given that the maximum network delay is known (e.g., within a single Wasp Committee that always maintains local connectivity or within a committee of nodes that uses wired connectivity, such as a network of charging stations). It is worth noting that this is a research network where functionality takes precedence over performance. In any case, the approach proposed in this paper is already valid for scenarios where consensus is not required in real-time but low impact on the use of computational resources is preferred.

In summary, we can conclude that distributed ledger technologies (DLTs) have potential to bring new standards of security and trust to large-scale robotic systems. Specific domains of interest include collaborative multi-robot systems and swarm robotic systems.

In addition, we have manifested that porting decision-making tasks on blockchain will not consume extra power compared to usual robotic tasks such as mapping. Despite all these benefits, there is a series of challenges that can be the objective of further investigation. First, there should be prior knowledge about the possible network splits since the disconnected nodes should still operate on the same chain. Second, this framework can only be applied if there are enough nodes in the network to form different chains. A third challenge is the limited capability of smart contracts in executing heavy decision-making tasks like machine learning based solutions. In future work, we are investigating how this kind of heavy process can be executed off-ledger. Fourth, based on the current implementations, we are working to report a thorough analysis of the performance of Wasp and GoShimmer nodes in robotic applications communicating with ROS\,2. Finally, we have noted that the current version of IOTA supporting smart contracts is a research network where functionality is sought out over performance. Theoretically, the network design implies a more scalable solutions. However, there is a limited amount of real-world deployments. In any case, our experiments prove the effectiveness for many applications that do not require fast consensus for the smart contracts within the Wasp committees, but where data is already fed in real-time to the GoShimmer layer.

\section{Conclusion}\label{sec:conclusion}

This paper proposes a framework for partition-tolerant decision-making processes in multi-robot systems. Smart contracts that incorporate the mentioned constraints are split into two smart contracts on IOTA's smart contract platform, which tolerates network partitioning. Furthermore, we demonstrate how IOTA's two-layer structure and ROS\,2 can be applied to a multi-robot system in a novel architectural design. In order to demonstrate the partition-tolerant framework and the proposed architecture, we chose a distributed mapping problem for a proof of concept. The distributed mapping task was first simulated in the Gazebo with eight robots. We then tested our distributed mapping smart contracts in a real-world scenario and compared the results to a baseline Ethereum implementation. Our results demonstrate how network partitioning impacts distributed decision-making outcomes in both cases, with only the IOTA-based implementation being able to showcase both byzantine-tolerant and partition-tolerant behaviour.

Future research will be directed towards more diverse applications and larger-scale experiments. Specifically, we will work on problems requiring more complex data processing, such as applications where neural networks are required for data processing but cannot be directly implemented within smart contracts.

\section*{Acknowledgment}

This research work is supported by the R3Swarms project funded by the Secure Systems Research Center (SSRC), Technology Innovation Institute (TII).

\bibliographystyle{unsrt}
\bibliography{bibliography}

\newpage
\begin{IEEEbiography}[{\includegraphics[width=0.94in,height=1.25in,clip,keepaspectratio]{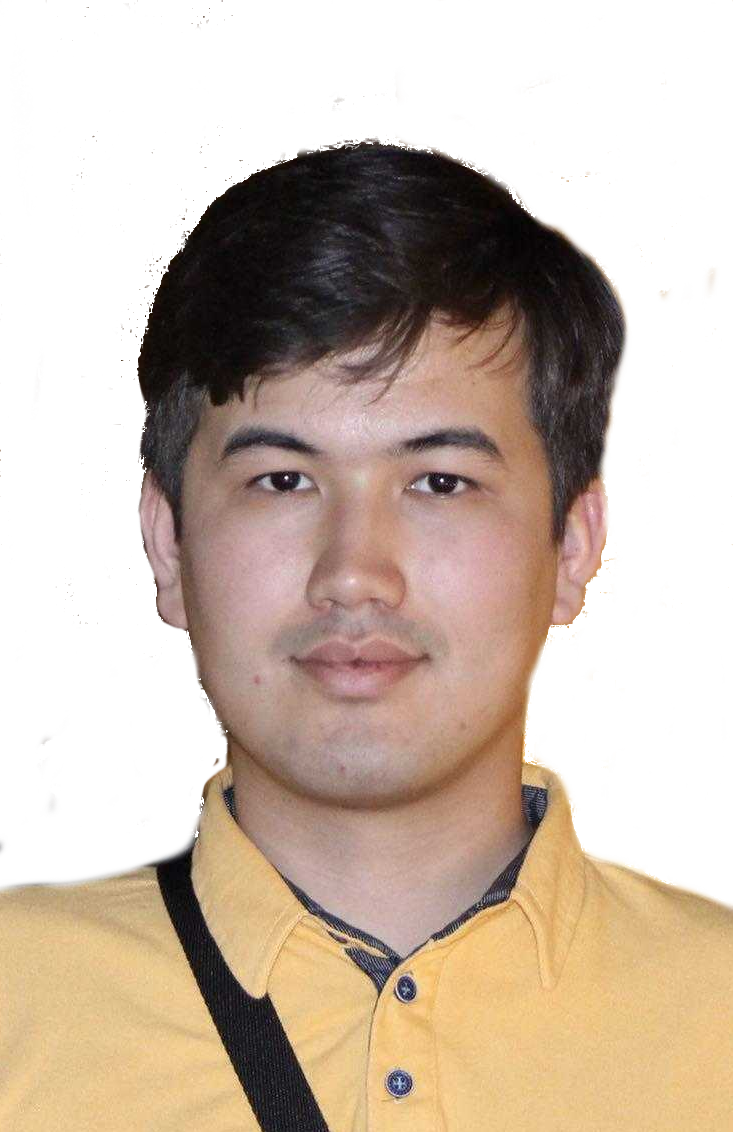}}]{Farhad Keramat} received his B.S. degree in Electrical Engineering and his M.Sc. degree in Secure Communication and Cryptography, from University of Tehran, Tehran, Iran, in 2017 and 2020, respectively. Since 2021, he has been a researcher at the Turku Intelligent Embedded and Robotic Systems (TIERS) Group, University of Turku. His research interests include distributed ledger technologies, multi-robot systems security and multi-robot collaboration.
\end{IEEEbiography}

\vspace{-1em}

\begin{IEEEbiography}[{\includegraphics[width=0.94in,height=1.25in,clip,keepaspectratio]{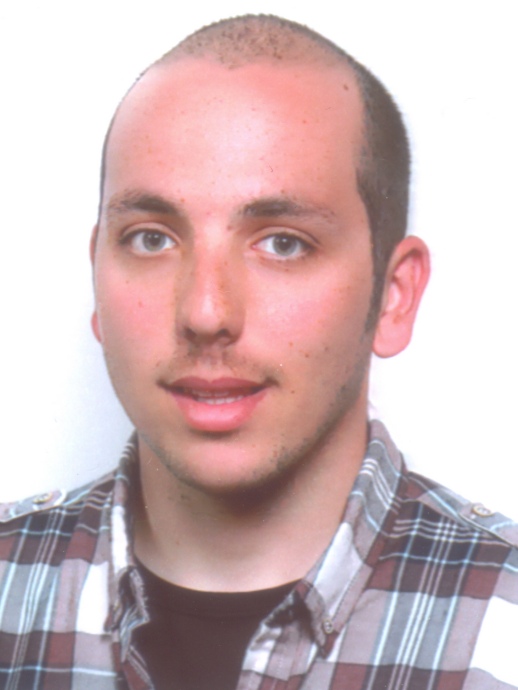}}]{Jorge Peña Queralta} received B.S. degrees in mathematics and physics engineering from UPC BarcelonaTech, Spain, in 2016, a M.Sc. (Tech.) degree in Information and Communication Science and Technology from the University of Turku, Finland, and a M. Eng. degree in Electronics and Communication Engineering from Fudan University, China, in 2018. Since 2018, he has been a researcher and doctoral candidate at the Turku Intelligent Embedded and Robotic Systems (TIERS) Group, University of Turku. His research interests include multi-robot systems, collaborative autonomy, distributed perception, and edge computing.
\end{IEEEbiography}

\vspace{-1em}

\begin{IEEEbiography}[{\includegraphics[width=1in,height=1.25in,clip,keepaspectratio]{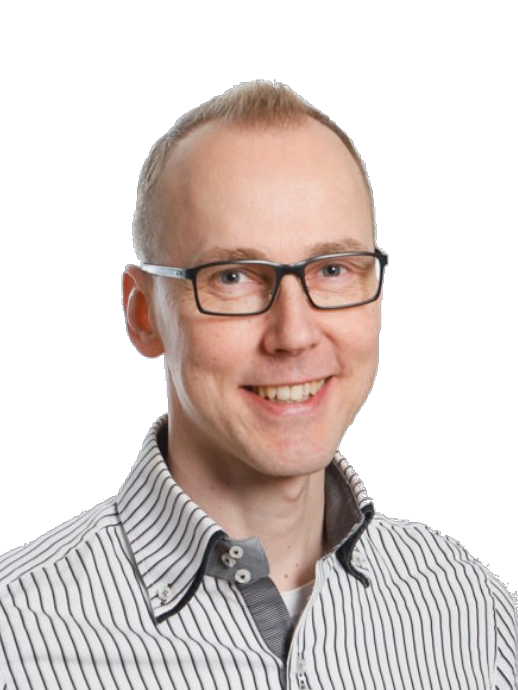}}]{Tomi Westerlund} is an Associate Professor of Autonomous Systems and Robotics at the University of Turku and a Research Professor at Wuxi Institute of Fudan University, Wuxi, China. Dr. Westerlund leads the Turku Intelligent Embedded and Robotic Systems research group (tiers.utu.fi), University of Turku, Finland. His current research interest is in the areas of Industrial IoT, smart cities and autonomous vehicles (aerial, ground and surface) as well as (co-)robots. In all these application areas, the core research interests are in multi-robot systems, collaborative sensing, interoperability, fog and edge computing, and edge AI.
\end{IEEEbiography}

\vspace*{\fill}

\end{document}